\documentclass{article}

 \usepackage[final]{nips_2017}

\usepackage[utf8]{inputenc} 
\usepackage[T1]{fontenc}    
\usepackage{hyperref}       
\usepackage{url}            
\usepackage{booktabs}       
\usepackage{amsfonts}       
\usepackage{nicefrac}       
\usepackage{microtype}      

\usepackage{graphicx}

\usepackage{amsmath}

\usepackage[subrefformat=parens]{subcaption}

\usepackage{wrapfig}

\title{Deep Residual Networks and Weight Initialization}

\author{
Masato Taki \\
  interdisciplinary Theoretical \& Mathematical Sciences Program (iTHEMS)\\
  RIKEN\\
  2-1, Hirosawa, Wako, Saitama 351-0198, Japan \\
  \texttt{taki@riken.jp} \\
}

\begin{document}

\maketitle

\begin{abstract}
Residual Network (ResNet) is the state-of-the-art architecture that realizes successful training of really deep neural network.
It is also known that good weight initialization of neural network avoids problem of vanishing/exploding gradients.
In this paper, 
simplified models of ResNets are analyzed.
We argue that goodness of ResNet is correlated with the fact
that ResNets are relatively insensitive to choice of initial weights.
We also demonstrate how  batch normalization improves  backpropagation of deep ResNets
without tuning initial values of weights.
\end{abstract}

\section{Introduction}

In the last few years,
developments in deep neural networks [1][2]
have derived
tremendous improvements in image recognition [3][4][5] and other machine learning tasks.
Increasing depth significantly yields accuracy gains,
but a deeper model suffers from serious problem of vanishing/exploding gradients [6][7] in general.
Initializing weight parameters by sampling them from an appropriate distribution is
a standard way nowadays to address this problem.
The first modern initialization method
was proposed in [8] for symmetric activation functions,
and it was widely applied to various architectures.
ReLU [9], which is not symmetric function,  becomes a popular selection of an activation function recently. 
He et al. [10] generalized the initialization [8] to such activation function.
So far, many researches have been made to extend this initialization method.

Residual network [11]\footnote{See also highway network [12].} is an another idea to realize deep neural networks with high performance.
Intuitive idea is that
shortcut connections in ResNet provide bypaths for propagating signals
and it prevents the loss of signal through propagation.

In this paper,
we provide another theoretical explanation of goodness of ResNet.
We show that ResNet
is relatively insensitive to weight initialization compared to usual neural networks.
This fact enable us to train deep ResNet successfully and easily.
ResNet is comparatively robust to initialization distribution,
but the variance of the initialization distribution must be smaller that that in [10] :
\begin{equation}
\textrm{Var}\big[ {w}\big]=\frac{c}{nL},
\end{equation}
where $L$ is the depth of ResNet\footnote{In this paper, this $L$ actually corresponds to
the numbers of residual blocks in ResNet.},
and $n$ is the fan-in. $c$ is some $\mathcal{O}(1)$ coefficient for instance.
For $L\gg1$, this requirement leads to a theoretical upper bound on possible value of $L$
when we realize it as practical floating-point arithmetic.
To adress this difficulty,
we investigate a well-known modification of ResNet with insertion of batch normalization layers [13].
Explosion of gradients  is then relaxed as
\begin{equation}
\textrm{Var}\left[ \frac{\partial E}{\partial w^\ell}\right]\propto \frac{L}{\ell}.
\end{equation}
This explosion is linear in the depth and not so serious as exponential explosion of plain networks [8][10].
Batch normalization significantly improves the problem,
but gradients still diverge when ResNet is extremely deep.
It can be an another theoretical limitation to possible depth of ResNet.

Deeply related work to this paper is Balduzzi et al. [23]:
similar analysis for special cases was done from the view point of shattered gradient problem.

\section{Weight Initialization for Residual Networks}

Careful weight initialization is the main current of the technique
for training deep neural networks without problem of vanishing/exploding gradients [14][15][16].
The behavior of a plain neural network is highly sensitive to the initial weights when the network is very deep,
and their distribution
directly affects the magnitudes of the outputs and gradients of the network [8][10].
We therefore need to tune the initial weight distribution carefully
for avoiding vanishing and explosion.

Using residual networks is another way to realize smooth convergence of training deep model.
Shortcut connections in ResNet 
keep
signals finite through propagation,
and this has been an intuitive explanation why ResNet can avoid problem of vanishing/exploding gradients.

In this section, we point out that these two approaches are actually related
and deep ResNet is special model from the viewpoint of the weight initialization.
We also propose new weight initialization that works for deep ResNets.

\subsection{Forward propagation}

We start with generalizing known initializations to the case of ResNets.
In this paper,
$\boldsymbol{W}^\ell=(w^\ell_{ji})$,
 $\boldsymbol{u}^\ell$
and $\boldsymbol{z}^\ell$ denote
the weight matrix between $(\ell-1)$-th and $\ell$-th layers,
the activation and the layer output
respectively.
Our toy model [17][18] of ResNet, whose block is illustrated in Fig.1a,
is then defined by the following feedforward propagation rule\footnote{We omit activation function after the addition of two signals
because effect of such activation is minor [19][20].} 
\begin{equation}
\label{feedforward}
\boldsymbol{u}^\ell = \boldsymbol{W}^\ell \boldsymbol{z}^{\ell-1},\quad
\boldsymbol{z}^\ell = f\big( \boldsymbol{u}^{\ell} \big) + \boldsymbol{z}^{\ell-1},
\quad (\ell=1,2,\cdots, L).
\end{equation}
We use a common activation function $f$ for all residual layers.
$\boldsymbol{x}=\boldsymbol{z}^0$ is the input, and $\boldsymbol{y}=\boldsymbol{z}^L$
is the output of the model.
We assume that all weights $w^\ell_{ji}$ are {\it{i.i.d.}} and sampled from a symmetric distribution $P(w^\ell)=P(-w^\ell)$.
Because of this symmetry, the activation ${u}^\ell$ also forms a symmetric distribution $P(u^\ell)=P(-u^\ell)$,
and we find
$
\textrm{E}\big[ {u}^\ell\big] =\textrm{E}\big[ -{u}^\ell\big]=0$.
We also have $\textrm{E}\big[ {w}^\ell\big]=0$.

To see problem of vanishing/exploding gradient,
we study decay/growth of the variances of the layer outputs $z^\ell_j$.
Assuming ${w}^\ell$ and ${z}^{\ell-1}$ are independent, the variance of the activation can be recast into
\begin{equation}
\label{uVSz}
\textrm{Var}\big[ {u}^\ell\big] 
= n\, \textrm{Var}\big[ {w}^\ell\big]
\textrm{E}\big[ \big({z}^{\ell-1}\big)^2 \big]
=
 n\, \textrm{Var}\big[ {w}^\ell\big]
\textrm{Var} \big[ {z}^{\ell-1} \big]
 +
n\, \textrm{Var}\big[ {w}^\ell\big]
\big(\textrm{E}\big[ {z}^{\ell-1} \big]\big)^2,
\end{equation}
where $n$ is the fan-in that is independent of $\ell$ in our model.
The first equality follows from $\textrm{E}\big[w^{\ell} \big]=0$.
Evaluating the expectation value $ \textrm{E}\big[ \big({z}^{\ell}\big)^2 \big]$ is next step.
The feedforward propagation rule (\ref{feedforward}) immediately leads to the recursion
\begin{equation}
\label{expZ2}
\textrm{E}\big[ \big({z}^\ell\big)^2\big] 
= \textrm{E}\big[ \big(f({u}^\ell)+ {z}^{\ell-1} \big)^2\big]
=  \textrm{E}\big[ \big(f({u}^\ell) \big)^2\big]
+ 2\,\textrm{E}\big[ f({u}^\ell) {z}^{\ell-1} \big]
+ \textrm{E}\big[ \big( {z}^{\ell-1} \big)^2\big].
\end{equation}
To solve this recursion relation, we beed to evaluate the layer output $f(u^\ell)$.
The behavior of it depends on the form of $f$,
and therefore, we divide the following discussion into following two cases.

\subsubsection*{Identity}
Symmetric activation functions $f(-u)=-f(u)$, such as $\tanh$, have been widely used.
For such activation functions, the second term of the right hand side of (\ref{expZ2}) vanishes automatically
$\textrm{E}\big[ f\big({u}^\ell \big)\big] \textrm{E}\big[{z}^{\ell-1} \big]=\textrm{E}\big[-f\big({u}^\ell \big)\big] \textrm{E}\big[{z}^{\ell-1} \big]=0$.
This is because that  $u^\ell$ and $z^{\ell-1}$ are independent random variables in our setup\footnote{
It is easy to see this fact. For instance, we can show
$
\textrm{Cov}\big[ u_i^\ell, z_j^{\ell-1}\big]
=
\textrm{E}\big[ u_i^\ell\big(z_j^{\ell-1}-\textrm{E}[z_j^{\ell-1}]\big)\big]
=\sum_k \textrm{E}\big[ w^\ell_{ik}z^{\ell-1}_k\big(z_j^{\ell-1}-\textrm{E}[z_j^{\ell-1}]\big)\big]
=0$
by using $\textrm{E}\big[w^{\ell} \big]=0$.},
and the second term of the right hand side of (\ref{expZ2}) is
$\textrm{E}\big[ f({u}^\ell) {z}^{\ell-1} \big]
=\textrm{E}\big[ f({u}^\ell) \big] \textrm{E}\big[ {z}^{\ell-1} \big]$.
Following [8], we approximate this activation function with identity map.
The equation (\ref{expZ2}) then leads to the following simple recursion
\begin{equation}
\textrm{E}\big[ \big({z}^\ell\big)^2\big] 
=  \textrm{E}\big[ \big({u}^\ell \big)^2\big]
+ \textrm{E}\big[ \big( {z}^{\ell-1} \big)^2\big]
=
\left(1+n\, \textrm{Var}\big[ {w}^\ell\big]\right)
 \textrm{E}\big[ \big({z}^{\ell-1}\big)^2 \big].
\end{equation}
We can therefore solve this relation easily,
and the expectation value at the output layer is given by
\begin{equation}
\label{formula1}
\textrm{E}\big[ ({z}^\ell)^2\big] 
= \textrm{E}\big[ {x}^2 \big]
\prod_{\ell'=1}^\ell
\left(1+n\, \textrm{Var}\big[ {w}^{\ell'}\big]\right).
\end{equation}
When $f$ is symmetric function,
we can easily show
$
\textrm{E}\big[ {z}^\ell\big] 
=
\textrm{E}\big[ f({u}^\ell)\big] 
+
\textrm{E}\big[ {z}^{\ell-1}\big] 
=
\textrm{E}\big[ {z}^{\ell-1}\big] 
=
\textrm{E}\big[ {x}\big]$.
Asumming that the input is normalized as $\textrm{E}\big[ {x}\big]=0$\footnote{This assumption
is only for simplicity. We can generalize our argument to generic case.} leads to
$\textrm{E}\big[ {z}^\ell\big]=0$ for all layers $\ell = 0,1,\cdots,L$.
The evolution equation for the variance of $z^\ell$ is therefor given by
\begin{equation}
\label{explosionId}
\textrm{Var}\big[ {z}^\ell\big] 
= \textrm{Var}\big[ {x} \big]
\prod_{\ell'=1}^\ell
\left(1+n\, \textrm{Var}\big[ {w}^{\ell'}\big]\right).
\end{equation}
Let us assume the distribution $P(w^\ell)$ is independent of $\ell$ for simplicity.
The prefactor for $\textrm{Var}[z^L]$ is then $\prod_{\ell=1}^L
\left(1+n\, \textrm{Var}\big[ {w}^\ell\big]\right)=\left(1+n\, \textrm{Var}\big[ {w}\big]\right)^L$,
and we want to keep this factor finite---$\mathcal{O}(1)$ for instance---to avoid explosion.
For very deep model $L\gg1$,
the simple choice to realize it is
\begin{equation}
\left(1+n\, \textrm{Var}\big[ {w}\big]\right)^L
=e^{L\log \left(1+n\, \textrm{Var}[ {w}]\right)}
\approx
e^{c},
\end{equation}
by choosing weight initialization distribution whose variance is
\begin{equation}
\label{recommend2}
\textrm{Var}\big[ {w}\big]=\frac{c}{nL}.
\end{equation}
Here $c$ is some small number.
This weight initialization distribution relies not only on local layer information $n$\footnote{
For sparse network such as convolutional network, 
the fan-in $n$ is not the actual number of the units.}
but also on global information---the depth $d$.

The relevant difference with plain feedforward neural networks [8] is 
that the prefactor in that case behaves
as $
\left(n\, \textrm{Var}\big[ {w}\big]\right)^L
$,
and therefore it is highly sensitive to deviation from the recommended value $n\, \textrm{Var}\big[ {w}\big]=1$.
If the variance $\textrm{Var}\big[ {w}\big]$ is twice the recommended value,
then the prefactor grows to $2^L$.
This number is extremely large for deep model: $2^{20}\approx 1000000$ for $L=20$.
Residual case (\ref{recommend2}), on the other hand, is more robust to such deviation.
If $c=1$ the grown prefactor is just
$
e^2\approx 7$.
This robustness eases training of very deep ResNet under our initialization.

\begin{figure}[t]
\centering
\begin{tabular}[t]{cc}
 \includegraphics[scale=0.5]{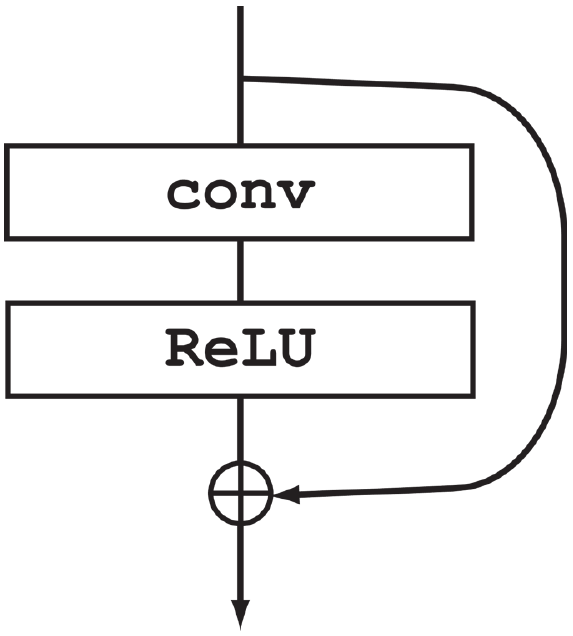}
 &
 \includegraphics[scale=0.5]{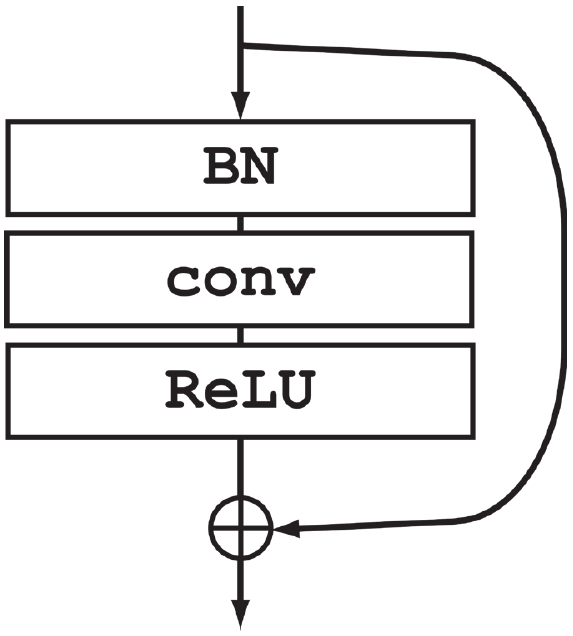}
\\
{(a) plain residual block}
 &
{(b) residual block with batch normalization}
 \end{tabular}
 \caption{Residual blocks. Here we use convolutional layer and ReLU for example.}\label{resblocks}
\end{figure}

\subsubsection*{ReLU}
The ReLU activation function $f(u)=\max (0, u)$ is another standard choice.
Since the distribution of $u^\ell$ is symmetric $P(u^\ell)=P(-u^\ell)$,
the expectation value of $f(u)$ and $\big(f(u)\big)^2$ are
\begin{equation}
\textrm{E}\big[ f({u}^\ell) \big]=\frac{1}{2}\,\textrm{E}\big[ \vert{u}^\ell\vert\big],
\quad
\textrm{E}\big[ \big(f({u}^\ell) \big)^2\big]=\frac{1}{2}\,\textrm{E}\big[ \big({u}^\ell\big)^2\big]
=\frac{1}{2}\,\textrm{Var}\big[ {u}^\ell\big].
\end{equation}

Let us subtract $\big(\textrm{E}\big[ {z}^\ell\big] \big)^2
=
\big(\textrm{E}\big[ f({u}^\ell)\big] \big)^2
+
2 \textrm{E}\big[ f({u}^\ell)\big]\textrm{E}\big[ {z}^{\ell-1}\big] 
+
\big(\textrm{E}\big[ {z}^{\ell-1}\big] \big)^2$ from (\ref{expZ2}) to calculate the variance.
The variance of $z^\ell$ is then
\begin{eqnarray}
&&\textrm{Var}\big[ {z}^\ell\big] 
=
\frac{1}{2}\textrm{Var}\big[ {u}^\ell\big] 
+
\textrm{Var}\big[ {z}^{\ell-1}\big] 
-\frac{1}{4}
\big(\textrm{E}\big[ \vert {u}^{\ell-1}\vert \big]\big)^2
\nonumber\\
&&=\left(1+\frac{1}{2}n\, \textrm{Var}\big[ {w}^\ell\big]\right)
\textrm{Var}\big[ {z}^{\ell-1} \big]
+\frac{1}{2}n\, \textrm{Var}\big[ {w}^\ell\big]
\big(\textrm{E}\big[ {z}^{\ell-1} \big]\big)^2
-\frac{1}{4}\big(\textrm{E}\big[ \vert{u}^\ell\vert\big]\big)^2.
\end{eqnarray}
We use (\ref{uVSz}) to obtain the last equality.

Because solving this recursion relation is hard, we evaluate a lower bound of $\textrm{Var}[z^\ell]$.
Jensen's inequality leads to the inequality
$\big(\textrm{E}\big[ \vert{u}^\ell\vert\big]\big)^2
\leq
\textrm{E}\big[ \big({u}^\ell\big)^2\big]
=n \textrm{Var}\big[ {w}^{\ell}\big]
\textrm{E}\big[ \big({z}^{\ell-1}\big)^2\big]
$,
and it proves the following lower bound of the variance
\begin{eqnarray}
\textrm{Var}\big[ {z}^\ell\big]
&\geq&
\left(1+\frac{1}{4}n\, \textrm{Var}\big[ {w}^\ell\big]\right)
\textrm{Var}\big[ {z}^{\ell-1} \big]
+\frac{1}{4}n\, \textrm{Var}\big[ {w}^\ell\big]\big(\textrm{E}\big[ {z}^{\ell-1} \big]\big)^2
\nonumber\\
&\geq&
\left(1+\frac{1}{4}n\, \textrm{Var}\big[ {w}^\ell\big]\right)
\textrm{Var}\big[ {z}^{\ell-1} \big].
\end{eqnarray}
The variance of layer output $z^\ell$ is therefore bounded below by the following value
\begin{eqnarray}
\label{explosionReLU}
\textrm{Var}\big[ z^\ell \big]
\geq
\textrm{Var}\big[ x \big]
\left(1+\frac{1}{4}n\, \textrm{Var}\big[ {w}\big]\right)^\ell.
\end{eqnarray}
The structure of this lower bound is the same as that of (\ref{formula1}),
and initialization distribution which is necessary condition for
avoiding this divergence
is again $
\textrm{Var}\big[ {w}\big]
={c}/{nL}$.
This condition guarantees $\textrm{Var}\big[ y \big]\geq e^{\frac{c}{4}} \,\textrm{Var}\big[ x \big]$.

\subsection{Back propagation}

Let us move on to evaluating decay/growth of gradients of ResNet.
In backpropagation method,
the gradients are given by the deltas as
\begin{equation}
\label{defdelta}
\frac{\partial E}{\partial \boldsymbol{W}^\ell}
=
\boldsymbol{\delta}^\ell\otimes
\boldsymbol{z}^{\ell-1},\quad
\boldsymbol{\delta}^\ell
=
\frac{\partial E}{\partial \boldsymbol{u}^\ell}
=f'(\boldsymbol{u}^{\ell})\odot 
\boldsymbol{\delta}^{\ell}_{{z}},\quad
\boldsymbol{\delta}^\ell_{{z}}
=
\frac{\partial E}{\partial \boldsymbol{z}^\ell},
\end{equation}
where $\otimes$ is the Kronecker product
and
$\odot$ is the Hadamard product.
The chain rule then leads to the following recursion relation for deltas between adjoining layers
\begin{equation}
\label{deltarecursionReLU}
\boldsymbol{\delta}^\ell_{{z}}
=
\boldsymbol{\delta}^{\ell+1}_{{z}}
+
\big(\boldsymbol{W}^{\ell+1}\big)^\top\left(
f'(\boldsymbol{u}^{\ell+1})\odot 
\boldsymbol{\delta}^{\ell+1}_{{z}}\right).
\end{equation}
Solving this recursion starting from the initial value $\boldsymbol{\delta}^L_{{z}}
=
{\partial E}/{\partial \boldsymbol{y}}$ and substituting the solution to (\ref{defdelta})
give the gradients for all layers.

\subsubsection*{ReLU}
Let us start with the case of the ReLU activation function.
To evaluate the variance of gradients,
we  assume that $f'(\boldsymbol{u}^{\ell})$ and $\boldsymbol{\delta}^\ell_{{z}}$ are also independent
random variables and each of them is {\it{i.i.d.}}.
The expectation value of the deltas are then
\begin{equation}
\textrm{E}\big[ {\delta}^\ell_{{z}} \big]
=
\textrm{E}\big[ {\delta}^{\ell+1}_{{z}} \big]
+
\sum_k
\textrm{E}\big[ \big({w}^{\ell+1}_{kj}\big)\big]\,
\textrm{E}\big[ 
f'({u}^{\ell+1}_k) 
{\delta}^{\ell+1}_{z,k}\big]
=\textrm{E}\big[ {\delta}^{\ell+1}_{{z}} \big],
\end{equation}
and $f'(u)$ becomes $0$ with probability $1/2$ and $1$ with probability $1/2$.
The relation (\ref{defdelta}) then gives
\begin{equation}
\textrm{E}\big[ {\delta}^\ell \big]
=
\textrm{E}\big[ 
f'({u}^{\ell}) \big]
\textrm{E}\big[ 
{\delta}^{\ell}_{z}\big]
=
\frac{1}{2}
\textrm{E}\big[ 
{\delta}^{\ell}_{z}\big]=
\frac{1}{2}
\textrm{E}\big[ 
{\delta}^{\ell+1}_{z}\big]=\cdots=\frac{1}{2}
\textrm{E}\big[ 
{\delta}^{L}_{z}\big].
\end{equation}
For widely-used choices of output layer and error function,
the delta at the output layer takes the form
$
{\delta}^{L}_{z}
=
\frac{1}{N}\sum_{n\in\mathcal{D}}\left(
z^L(\boldsymbol{x}_n)-y_n
\right)$,
where
$\mathcal{D}$ is the training set.
The expectation value over models is
$
\textrm{E}\big[ {\delta}^{L}_{z}\big]
=
\textrm{E}\big[ \textrm{E}_{\mathcal{D}}[ 
z^L(x)]\big]
-
\textrm{E}_{\mathcal{D}}\big[ y\big]$.
It is natural that the output of the dataset and models are not biased,
and we can assume that $z^L(x)$ and $y$ take the same expectation value.
This assumption implies  $\textrm{E}\big[ {\delta}^\ell \big]=0$ for all layers.

Moreover the  relation (\ref{defdelta}) implies
$\textrm{E}\big[ ({\delta}^\ell)^2 \big]
=\textrm{E}\big[ (f'(u^\ell))^2 \big]\textrm{E}\big[ ({\delta}^\ell_z)^2 \big]
=\textrm{E}\big[ ({\delta}^\ell_z)^2 \big]/2
$,
and then the variance of the delta is 
\begin{equation}
\textrm{Var}\big[ {\delta}^\ell \big]
=\textrm{E}\big[ ({\delta}^\ell)^2 \big]-\frac{1}{4}\textrm{E}\big[ 
{\delta}^{\ell}_{z}\big]
=\frac{1}{2}\textrm{E}\big[ ({\delta}_z^\ell)^2 \big]
.
\end{equation}
Since the expectation value of $W^\ell$ is zero,
the variance in the last term becomes
\begin{equation}
\label{RNvariance1}
\textrm{E}\big[ ({\delta}_z^\ell)^2 \big]
=
\textrm{E}\big[ ({\delta}_z^{\ell+1})^2 \big]
+
\textrm{E}\bigg[ 
 \bigg({w}^{\ell+1}_{kj}
f'({u}^{\ell+1}_k) 
{\delta}^{\ell+1}_{z,k}\bigg)^2
\bigg]
=\left(
1+\frac{1}{2}n\,\textrm{Var}[w^{\ell+1}]
\right)\textrm{E}\big[ ({\delta}_z^{\ell+1})^2 \big].
\end{equation}
We use (\ref{deltarecursionReLU}) to derive the first equality.
Hence we obtain the following expression
\begin{equation}
\textrm{Var}\big[ {\delta}^\ell \big]
=\frac{1}{2}
\textrm{Var}\big[ {\delta}^L_z \big]
\prod_{\ell'=\ell}^{L-1}
\left(
1+\frac{1}{2}n\,\textrm{Var}[w^{\ell'+1}]
\right).
\end{equation}
Using common initialization distribution for all layers,
we get a simplified formula
$
\textrm{Var}\big[ {\delta}^\ell \big]
=
\textrm{Var}\big[ {\delta}^L_z \big]
\left(
1+n\,\textrm{Var}[w]/2
\right)^{L-\ell}/2$.
Notice that this factor
is similar to that of the formula (\ref{explosionReLU}).
Combining these two formulas then yields the lower bound of the variance of gradient
\begin{equation}
\textrm{Var}\left[ \frac{\partial E}{\partial w^\ell}\right]
>\frac{1}{4}
\textrm{Var}\big[ {\delta}^L_z \big]\textrm{Var}\big[ x \big]
\left(
1+\frac{1}{4}n\,\textrm{Var}[w]
\right)^{L-1}.
\end{equation}
We can conclude that the most simple choice of the initialization is again $\textrm{Var}\big[ {w}\big]
={c}/{nL}$ .

\subsubsection*{Identity}

When the activation function is approximated by the identity map $f(u)=u$,
the derivative of it is of course $f'(u)=1$.
This simplification leads to
\begin{equation}
\textrm{E}\big[ {\delta}^\ell \big]
=
\textrm{E}\big[ {\delta}_z^\ell \big],\quad
\textrm{E}\big[ {\delta}_z^\ell \big]
=\textrm{E}\big[ {\delta}_z^{\ell+1} \big]=\cdots
=\textrm{E}\big[ {\delta}_z^L \big].
\end{equation}
Let us assume $\textrm{E}\big[ {\delta}_z^L \big]=0$ again.
The variances of deltas then satisfy
\begin{equation}
\textrm{Var}\big[ {\delta}^\ell \big]
=
\textrm{E}\big[ ({\delta}_z^\ell)^2 \big].
\end{equation}

Using (\ref{deltarecursionReLU}), we obtain the recursion relation of this variance
\begin{equation}
\label{RNvariance2}
\textrm{E}\big[ ({\delta}_z^\ell)^2 \big]
=
\textrm{E}\big[ ({\delta}_z^{\ell+1})^2 \big]
+
\textrm{E}\bigg[ 
 \bigg({w}^{\ell+1}_{kj}
{\delta}^{\ell+1}_{z,k}\bigg)^2
\bigg]
=\bigg(
1+n\,\textrm{Var}[w^{\ell+1}]
\bigg)\textrm{E}\big[ ({\delta}_z^{\ell+1})^2 \big].
\end{equation}
This relation immediately leads to the following formula
\begin{equation}
\textrm{Var}\big[ {\delta}^\ell \big]
=
\textrm{Var}\big[ {\delta}^L_z \big]
\bigg(
1+n\,\textrm{Var}[w]
\bigg)^{L-\ell}.
\end{equation}
Since this growth factor is the same as (\ref{explosionId}),
we obtain the formula for the variance of gradients $\partial E/\partial w^\ell
=\delta^\ell z^{\ell-1}$
\begin{equation}
\textrm{Var}\left[ \frac{\partial E}{\partial w^\ell}\right]
=
\textrm{Var}\big[ {\delta}^L_z \big]\textrm{Var}\big[ x \big]
\left(
1+\frac{1}{2}n\,\textrm{Var}[w]
\right)^{L-1}.
\end{equation}
Therefore, this growth factor requires the same initialization distribution as the ReLU case.

\subsection{Limitation of depth}

In the previous subsection,
we propose new weight initialization distribution with the variance
\begin{equation}
\label{proposal}
\textrm{Var}\big[ {w}\big]=\frac{c}{nL},
\end{equation}
which is required to prevent exploding gradients.
Theoretically we can always prepare such initial weights for any $n$ and $L$,
but of course possible mantissa and exponent of practical float numbers are limited.
Implementation of deep learning sometimes prefers small bit-width
to save computational cost.
Using very small weight values causes cancellation and loss of trailing digits,
and thus there is an obstruction to realize variance
$\textrm{Var}\big[ {w}\big]
={c}/{nL}$ for very deep network.
This fact restricts possible depth of ResNet when
we implement our initialization as floating-point arithmetic.
In the next section, we compare our initialization
with another improvement known as batch normalization.

\section{Batch Normalization}
In this section,
we extend our previous analysis to the cases of batch normalized ResNets.
It is experimentally known that the introduction of batch normalization layers
improves performance of ResNets drastically.
This result, however, highly depends on the way to insert these layers,
and therefore we pick up a typical way.
Our theoretical treatment in this section explains how batch normalization
cures problems inhered in ResNets.

We consider batch normalization inserted before residual connection  as illustrated Fig.1b.
The feedforward propagation rule for a $\ell$-th residual block is then
\begin{equation}
\label{BAfeed}
\hat{\boldsymbol{z}}^{\ell-1}
=\frac{\boldsymbol{z}^{\ell-1} - \boldsymbol{\mu}^{\ell-1}}{\boldsymbol{\sigma}^{\ell-1}},\quad
\boldsymbol{u}^\ell=\boldsymbol{W}^\ell\hat{\boldsymbol{z}}^{\ell-1},\quad
\boldsymbol{z}^\ell=f(\boldsymbol{u}^\ell)+\boldsymbol{z}^{\ell-1}.
\end{equation}
We assume that batch size $\vert \mathcal{D}\vert$ is enough large
and the mean over the batch can be identified with the expectation value over the data generating distribution as
$
\boldsymbol{\mu}^{\ell}
=\textrm{E}\big[ \boldsymbol{z}^{\ell} \big]$
and
$\boldsymbol{\sigma}^{\ell}
=\textrm{E}\big[ (\boldsymbol{z}^{\ell})^2 \big]
-
(\boldsymbol{\mu}^{\ell})^2$.

Feedforward propagation has very simple property.
Let us focus on identity activation function for simplicity.
The expectation value of layer output is 
\begin{eqnarray}
&&\textrm{E}[z^\ell]
=\textrm{E}[f(u^\ell)]+\textrm{E}[z^{\ell-1}]
=\textrm{E}[z^{\ell-1}]=\cdots=\textrm{E}[x]=0,\\
&&\textrm{Var}[(z^{\ell})^2]
=n\textrm{Var}[w]+\textrm{Var}[(z^{\ell-1})^2]=\cdots
=n\ell \textrm{Var}[w] +\textrm{Var}[x]\approx
n\ell \textrm{Var}[w] 
\label{BNvarz}.
\end{eqnarray}

The derivation of the following backpropagation relation is parallel to that in previous section
\begin{equation}
\label{AAfeed}
\frac{\partial E}{\partial \boldsymbol{W}^\ell}
=
\boldsymbol{\delta}^\ell\otimes
\boldsymbol{\hat{z}}^{\ell-1},\quad
\boldsymbol{\delta}^\ell
=f'(\boldsymbol{u}^{\ell})\odot 
\boldsymbol{\delta}^{\ell}_{{z}}.
\end{equation}
The remaining relation we need to derive is that connecting $\boldsymbol{\delta}^\ell_{z}$ and $\boldsymbol{\delta}^{\ell+1}_{z}$.
With using chain rule of derivative we  obtain the backpropagation rule of deltas between adjoining layers
\begin{eqnarray}
\label{BCdeltaFormula0}
\delta^\ell_{z,j}
=
\sum_k
\frac{\partial E}{\partial {z}^{\ell+1}_{k}}
\frac{\partial {z}^{\ell+1}_{k}}{\partial z^\ell_j}
=\delta^{\ell+1}_{{z},j}
+\sum_{k,i}
f'(u^{\ell+1}_k)w^{\ell+1}_{ki}
\frac{\partial \hat{z}^\ell_{i}}{\partial z^\ell_j}
\delta^{\ell+1}_{{z},k}.
\end{eqnarray}
Chain rule of derivative again gives the derivative coefficient appearing in the second term
\begin{equation}
\label{ACdeltaFormula}
\frac{\partial \hat{z}^\ell_{i}}{\partial z^\ell_j}
=
\frac{1}{\sigma^\ell_i}\left(\delta_{i,j}-\frac{\partial \mu^\ell_i}{\partial z^\ell_j}\right)
-
\frac{1}{\left( \sigma^\ell_i \right)^2}
\frac{\partial \sigma^\ell_i}{\partial z^\ell_j}
\left(z^\ell_i-\mu^\ell_i\right)
.
\end{equation}
Here $\delta_{i,j}$ is the Kronecker delta,
and the mean and variance are originally those for $z^\ell$
over mini-batch samples
\begin{equation}
\mu^\ell_i
=
\frac{1}{N}
\sum_{n=1}^N
z^{\ell}_{n,i},
\quad
(\sigma^\ell_i)^2
=
\frac{1}{N}
\sum_{n=1}^N
(z^{\ell}_{n,i})^2-(\mu^\ell_i)^2,
\end{equation}
where $\boldsymbol{z}^{\ell}_{n}=( z^{\ell}_{n,i})$ is the layer output of the $n$-th sample in $\mathcal{D}$.
Then, the derivative coefficients appearing in (\ref{ACdeltaFormula}) are
\begin{equation}
\frac{\partial 
\mu^\ell_k}
{\partial z^{\ell}_{j}}
=
\frac{1}{N}
\delta_{k,j},
\quad
\frac{\partial 
\sigma^\ell_k}
{\partial z^{\ell}_{j}}
=
\frac{1}{N \sigma^\ell_k}
\delta_{k,j}
(z^{\ell}_{k}-\mu^\ell_k).
\end{equation}
Substituting these expressions into (\ref{ACdeltaFormula}),
we can rewrite the right hand side of (\ref{BCdeltaFormula0}) as
\begin{eqnarray}
\label{BCdeltaFormula}
\delta^\ell_{z,j}
=\delta^{\ell+1}_{{z},j}
+\sum_k
f'(u^{\ell+1}_k)w^{\ell+1}_{kj}\left(
\frac{1}{\sigma^\ell_j}\left(1-\frac{1}{N}\right)
-
\frac{1}{\left( \sigma^\ell_j \right)^3N}
\left(z^\ell_j-\mu^\ell_j\right)^2
\right)
\delta^{\ell+1}_{{z},k}.
\end{eqnarray}

We immediately get interesting property from this formula.
Because of $\textrm{E}[w^\ell]=0$,
the expectation value of the formula is
$
\label{BAmastereq}
\textrm{E}[\delta^\ell_{z,j}]
=
\textrm{E}[\delta^{\ell+1}_{{z},j}]$,
and all layers share the same value of the expectation of delta.
Our assumption $\textrm{E}\big[ {\delta}_z^L \big]=0$ then implies $\textrm{E}[\delta^\ell_{z}]=0$
for all layers.

Let us calculate the variance by using (\ref{BCdeltaFormula})
and the property $\textrm{E}[w^\ell_{kj}w^\ell_{k'j}]=\textrm{Var}[w]\delta_{kk'}$.
We first take the limit $N\to\infty$ to simplify expression.
Since $f'(u^{\ell+1})$, $w^{\ell+1}$ and $\delta^{\ell+1}_z$ are now independent random variables,
the expectation value of $(\delta^\ell_z)^2$ is
\begin{equation}
\label{BNmaster}
\textrm{Var}[
\delta^\ell_{z}]
=
\left(
1+
a\frac{n \textrm{Var}[w]}{(\sigma^\ell)^2}
\right)
\textrm{Var}[\delta^{\ell+1}_{z}],
\end{equation}
where $a$ is
\begin{equation}
a=\textrm{E}[f'(u^{\ell+1}_k)]=\begin{cases}
    1 & (identity) \\
    \frac{1}{2} & (ReLU)
  \end{cases}.
\end{equation}

To demonstrate how this formula avoids the explosion of the variance,
we focus on the case of identity activation function for simplisity.
The mean value appearing in the batch normalization is $
\mu^\ell=\textrm{E}[z^\ell]=\textrm{E}[u^\ell+z^{\ell-1}]=\textrm{E}[z^{\ell-1}]=\cdots=\textrm{E}[x]$,
and  it is therefore zero for all layers
since we can assume the input is normalized $\textrm{E}[x]=0$.
The variance over mini-batch is now identified with that over data generating distribution (\ref{BNvarz}),
and therefore we obtain
\begin{equation}
\label{sigmasolution}
(\sigma^\ell_j)^2
\approx
\ell n\textrm{Var}[w].
\end{equation}
Substituting it into (\ref{BNmaster}) leads to the following simple formula
\begin{equation}
\textrm{Var}[
\delta^\ell_{z}]
=
\textrm{Var}[\delta^{L}_{z}]
\prod_{\ell'=\ell}^{L-1}
\left(
1+
\frac{a}{\ell'}
\right),
\end{equation}
where $a=1$ for identity activation.
This formula makes the role of batch normalization clear.
Through the backpropagation,
the delta at the first layer blows up as 
$
\textrm{Var}[
\delta^\ell_{z}]
=
\textrm{Var}[\delta_{z}^L]
\prod_{\ell'=\ell}^{L-1}
\left(
1+
\frac{1}{\ell'}
\right)=\textrm{Var}[\delta_{z}^L]L/\ell
$.
Combining it with backpropagation rule then gives the following behavior
\begin{equation}
\textrm{Var}\left[ \frac{\partial E}{\partial w^\ell}\right]= \frac{L}{\ell}\,
\textrm{Var}[\delta_{z}^L]
.
\end{equation}

This linear growth\footnote{In Appendix, we study another batch normalization
and show the same property for identity and ReLU.
} is not small but still acceptable compared to exponential growth in plain ResNet.
This linear modification relaxes the explosion of gradient in deep ResNet.
Such linear growth of gradients
is not problematic in practical ResNets.
This is a mechanism how batch normalization improves the behavior of ResNet drastically.
But this factor for small $\ell$ causes serious explosion\footnote{We cannot improve it by tuning single learning rate
because the explosion factor depends on $\ell$.} when ResNet becomes extremely deep,
and then we cannot improve the explosion by weight initialization any more 
because there is no dependency on $\textrm{Var}[w]$ in this explosion\footnote{In (\ref{sigmasolution}),
we neglect $\textrm{Var}[x]$. This term could be relevant for small $\ell$,
and it might be possible to improve explosion by utilizing this term.}.
This this fact can be a drawback of batch normalization if one considers extremely deep situation.

\begin{figure}[t]
\centering
\begin{tabular}[t]{cc}
 \includegraphics[scale=0.3]{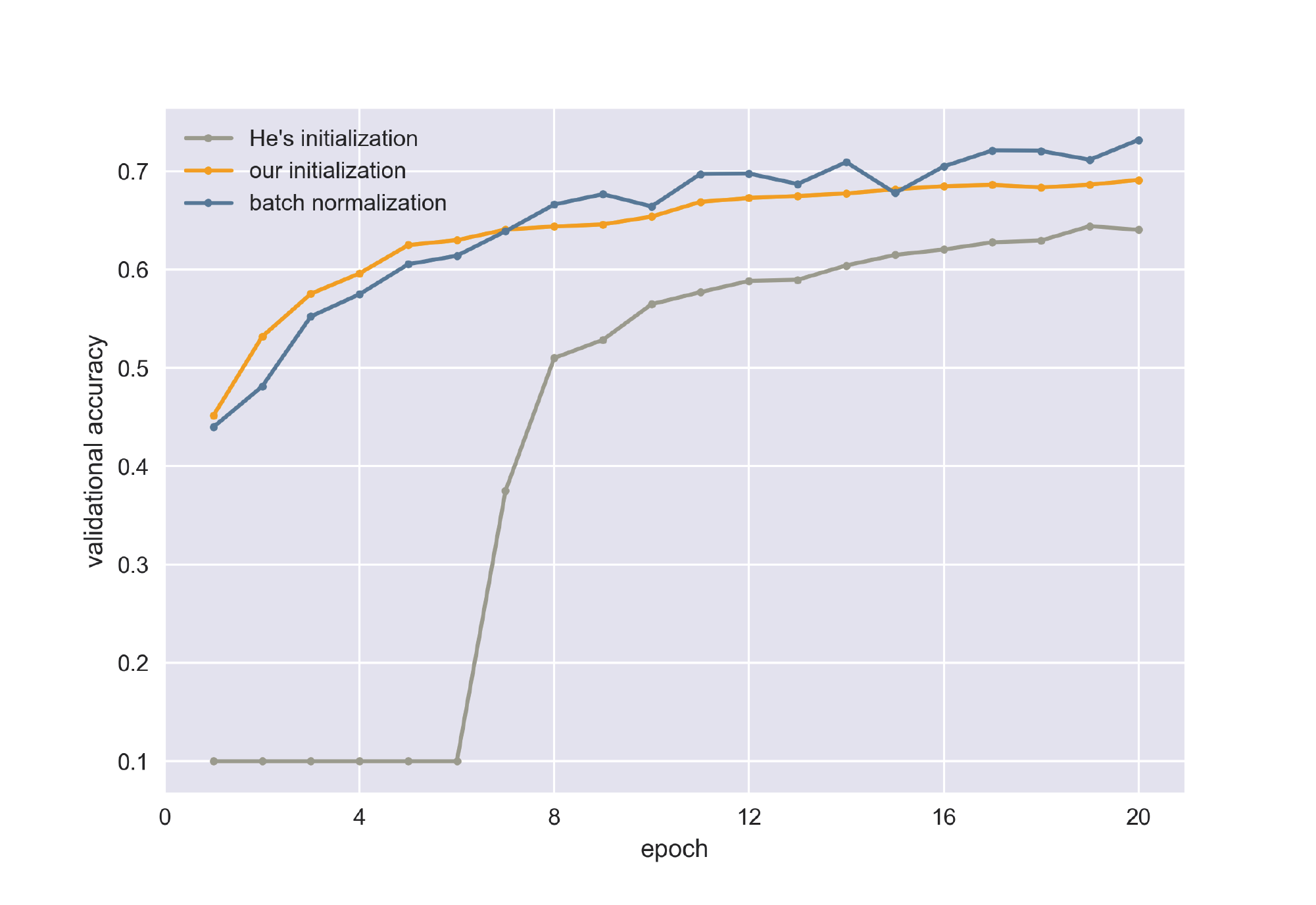}
 &
 \includegraphics[scale=0.3]{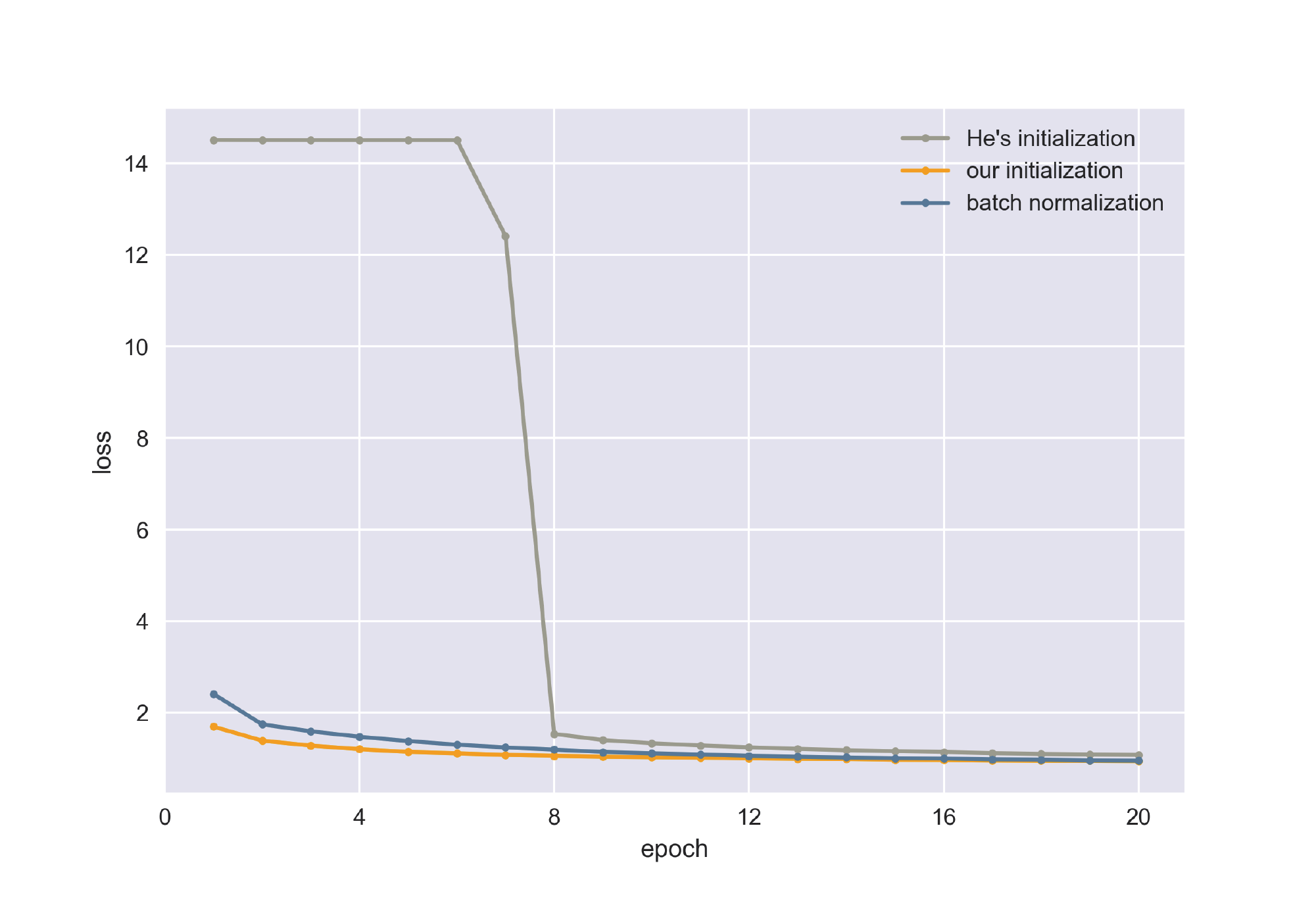}
\\
{(a) validation accuracy}
 &
{(b) loss}
 \end{tabular}
  \caption{Learning curves for ResNet with 100 residual blocks. 
  We compared three different setups.}\label{figuretwo}
\end{figure}

\section{Experiments}

To evaluate our initialization (\ref{proposal}),
we conduct little experiments on ResNets with a hundred of blocks.
Residual blocks without and with batch normalization layer
are illustrated in Fig.\ref{resblocks}.
Since our purpose is not realizing significant performance but checking effect of initializations
at the first stage of training,
we use simple models whose residual block consists of $8\times 8$ convolution layer with 16 channels
and ReLU activation function.
These models are trained on the CIFAR-10 dataset [21].

\begin{wrapfigure}[17]{r}[2mm]{70mm}
  \centering
  \includegraphics[keepaspectratio,width=70mm]
  {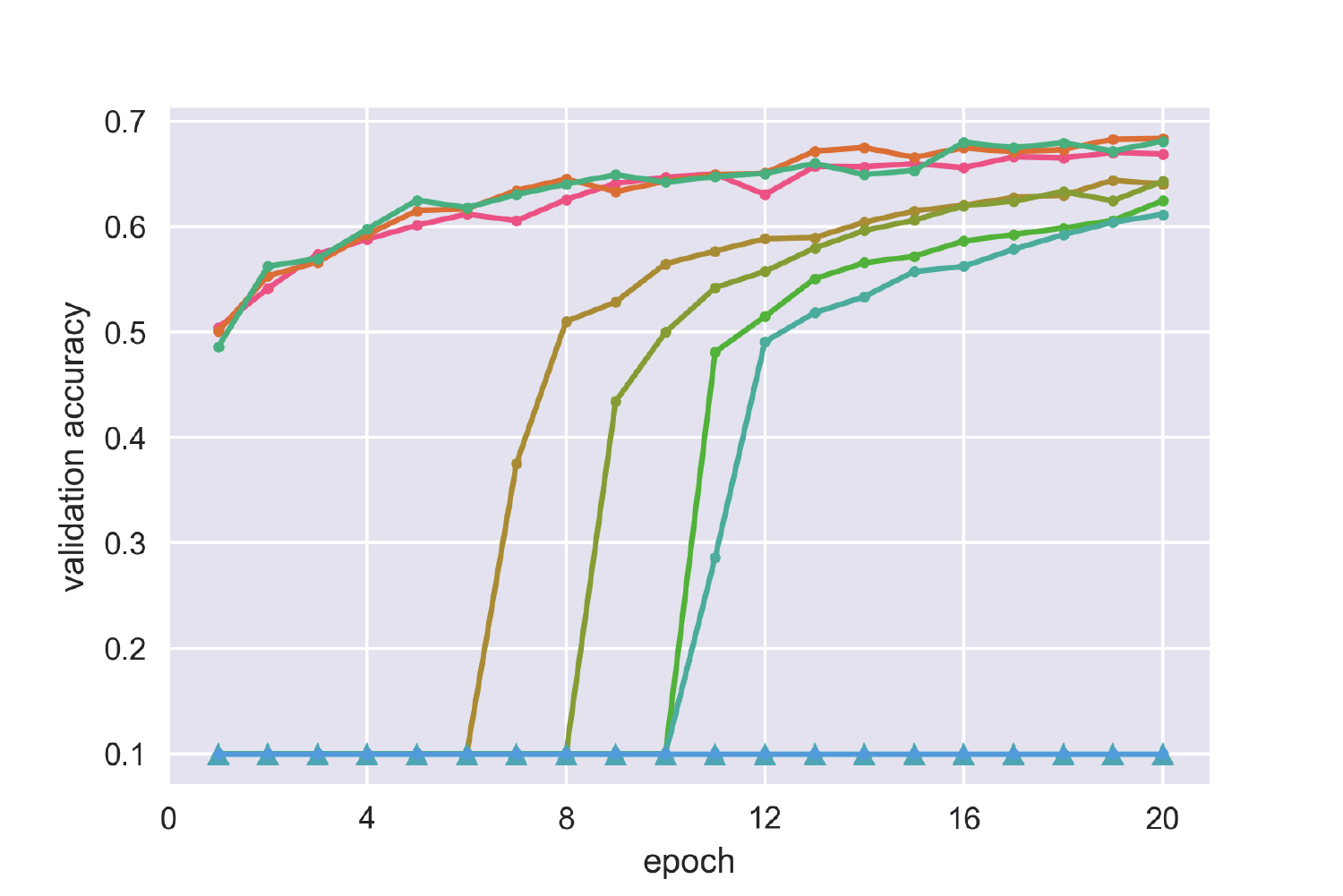}
  \caption{Nine independent learning curves of ResNet.
Weights are initialized by [10].}\label{figurethree}
\end{wrapfigure}
Learning curves at first 20 epochs are shown in Fig.\ref{figuretwo}.
Gray and yellow lines are learning curves of plain ResNets without batch normalization.
Gray line corresponds to He's initialization [10],
and yellow line corresponds to the proposed initialization (\ref{proposal})\footnote{We used
the normal distribution with variance $1/(8\times 8\times 16\times 100)$. This distribution in not truncated.}.
Learning curves with He's initialization under batch normalization are illustrated in blue lines.
Fig.2a is the validation accuracies,
and Fig.2b is the values of the loss functions at each epoch.

As Fig.\ref{figuretwo}, training of plain ResNet with He's initialization (gray lines) is
trapped on a plateau from the beginning and converge is hampered.
This initialization in this model is very unstable and easily trapped in plateau:
Fig.\ref{figurethree} shows nine learning curves of ResNets initialized by [10].
On the one hand, the same ResNet with our initialization (yellow lines)
is more stable: 
repeated experiments showed that the mean and variance of the validation accuracy after the first epoch are $0.434$ and $0.017$.
Our initialization in Fig.\ref{figuretwo} shows a comparable performance with ResNet with batch normalization (blue lines).
The performance of batch normalization is slightly better than our initialization,
and it stabilizes the convergence throughout the entire training period [22].
But batch normalization takes more computational cost than our simple initialization.
In this sense, our initialization can be substitute method to batch normalization.

\section{Discussion}
In this paper, we study how weight initialization and batch normalization improve
the training of ResNet. We also propose new weight initialization (\ref{proposal})
that works for deep ResNets. A simple experimental test shows that its effect at the first stage of training is comparable to that 
of batch normalization.

ResNets in this paper is more simplified than practical models,
and generalizing our analysis to complicated ResNets is an
interesting further direction.
It is especially important to understand the role of including activation function after shortcut connection,
inserting many layers in a single block
and changing the position of batch normalization layer [19].
We leave these questions open for future research.

\subsubsection*{Acknowledgments}

This work was supported by RIKEN interdisciplinary Theoretical \& Mathematical Sciences Program (iTHEMS)
and JSPS KAKENHI Grant Number JP17K19989.

\section*{References}

\small

[1] LeCun, Y., et al. (1989). Backpropagation applied to handwritten zip code recognition. {\it Neural computation}, 1(4), 541-551.

[2] Krizhevsky, A., Sutskever, I.,\ \&  Hinton, G. E. (2012). Imagenet classification with deep convolutional neural networks. {\it In Advances in neural information processing systems} (pp. 1097-1105).

[3] Russakovsky, Olga, et al. (2015). Imagenet large scale visual recognition challenge. {\it International Journal of Computer Vision}, 115(3), 211-252.

[4] Szegedy, C., et al. (2015). Going deeper with convolutions. In {\it Proceedings of the IEEE conference on computer vision and pattern recognition} (pp. 1-9).

[5] Simonyan, K.,\ \& Zisserman, A. (2014). Very deep convolutional networks for large-scale image recognition. {\it arXiv preprint} arXiv:1409.1556.

[6] S. Hochreiter. Untersuchungen zu dynamischen neuronalen Netzen. Diploma thesis, Institut fur Informatik, Lehrstuhl Prof. Brauer, Technische Universitat Munchen, 1991. 

[7] Bengio, Y., Simard, P.,\ \& Frasconi, P. (1994). Learning long-term dependencies with gradient descent is difficult. {\it IEEE transactions on neural networks}, 5(2), 157-166.

[8] Glorot, X.,\ \& Bengio, Y. (2010, March). Understanding the difficulty of training deep feedforward neural networks. In {\it Proceedings of the Thirteenth International Conference on Artificial Intelligence and Statistics} (pp. 249-256).

[9] Nair, V.,\ \& Hinton, G. E. (2010). Rectified linear units improve restricted boltzmann machines. In {\it Proceedings of the 27th international conference on machine learning} (ICML-10) (pp. 807-814).

[10] He, K., Zhang, X., Ren, S.,\ \& Sun, J. (2015). Delving deep into rectifiers: Surpassing human-level performance on imagenet classification. In {\it Proceedings of the IEEE international conference on computer vision} (pp. 1026-1034).

[11] He, K., Zhang, X., Ren, S.,\ \& Sun, J. (2016). Deep residual learning for image recognition. In {\it Proceedings of the IEEE conference on computer vision and pattern recognition} (pp. 770-778).

[12] Srivastava, R. K., Greff, K.,\ \& Schmidhuber, J. (2015). Highway networks. {\it  arXiv preprint} arXiv:1505.00387.

[13] Ioffe, S.,\ \& Szegedy, C. (2015, June). Batch normalization: Accelerating deep network training by reducing internal covariate shift. In {\it International Conference on Machine Learning} (pp. 448-456).

[14] LeCun, Y. A., et al. (2012). Efficient backprop. In {\it Neural networks: Tricks of the trade} (pp. 9-48). Springer Berlin Heidelberg.

[15] Saxe, A. M., McClelland, J. L.,\ \& Ganguli, S. (2013). Exact solutions to the nonlinear dynamics of learning in deep linear neural networks. {\it arXiv preprint }arXiv:1312.6120.

[16] Mishkin, D.,\ \& Matas, J. (2015). All you need is a good init. {\it arXiv preprint} arXiv:1511.06422.

[17] Veit, A., Wilber, M. J.,\ \& Belongie, S. (2016). Residual networks behave like ensembles of relatively shallow networks. In {\it Advances in Neural Information Processing Systems} (pp. 550-558).

[18] Greff, K., Srivastava, R. K.,\ \& Schmidhuber, J. (2016). Highway and residual networks learn unrolled iterative estimation. {\it arXiv preprint} arXiv:1612.07771.

[19] He, K., Zhang, X., Ren, S.,\ \& Sun, J. (2016, October). Identity mappings in deep residual networks. In {\it European Conference on Computer Vision} (pp. 630-645). Springer International Publishing.

[20] Gross, S.,\ \& M. Wilber. (2016) Training and investigating residual nets.\\
http://torch.ch/blog/2016/02/04/resnets.html 

[21] Krizhevsky, A.\ \& Hinton, G. (2009). Learning multiple layers of features from tiny images.

[22] Zhang, C., et al. (2016). Understanding deep learning requires rethinking generalization. {\it arXiv preprint} arXiv:1611.03530.

[23] Balduzzi, D., et al. (2017). The Shattered Gradients Problem: If resnets are the answer, then what is the question?. {\it arXiv preprint} arXiv:1702.08591.

\newpage
\subsubsection*{Appendix}

\begin{figure}[t]
\centering
 \includegraphics[scale=0.5]{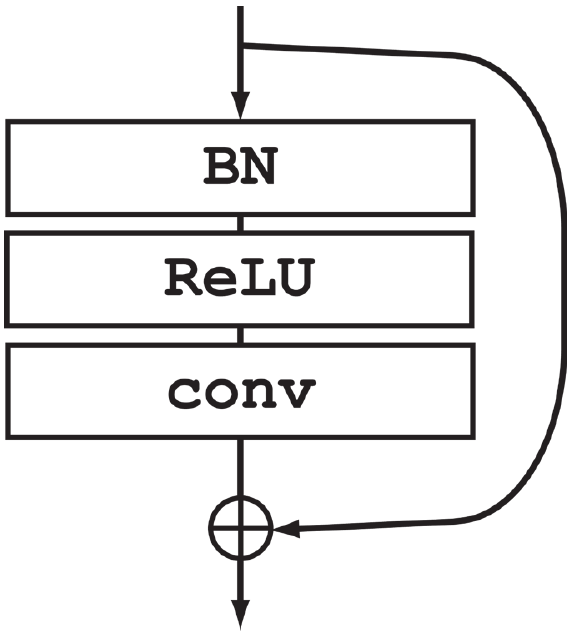}
 \caption{Residual block ReLU before convolution.}\label{appenfigure}
\end{figure}
In this appendix, we study another way of insertion of batch normalization layer Fig.\ref{appenfigure}.
In appendix of  [23], related computation is done for special case.

The feedforward propagation rule in this case Fig.\ref{appenfigure} is
\begin{equation}
\label{BAfeed2}
\hat{\boldsymbol{z}}^{\ell-1}
=\frac{\boldsymbol{z}^{\ell-1} - \boldsymbol{\mu}^{\ell-1}}{\boldsymbol{\sigma}^{\ell-1}},\quad
\boldsymbol{z}^\ell=\boldsymbol{W}^\ell f(\hat{\boldsymbol{z}}^{\ell-1})+\boldsymbol{z}^{\ell-1}.
\end{equation}
The expectation value of layer output is
\begin{equation}
\textrm{E}[
{z}^\ell]=\textrm{E}[{z}^{\ell-1}]
=\cdots
=\textrm{E}[x]=0.
\end{equation}
We therefore obtain the following expression for variance
\begin{equation}
\textrm{Var}[
{z}^\ell]=
n\textrm{Var}[
w]\,
\textrm{E}[
f(\hat{z}^\ell)^2]+
\textrm{Var}[{z}^{\ell-1}]
.
\end{equation}
The expectation value $\textrm{E}[
f(\hat{z}^\ell)^2]$ becomes
\begin{eqnarray}
\textrm{E}[
f(\hat{z}^\ell)^2]
=\begin{cases}
    \textrm{E}[
(\hat{z}^\ell)^2] & (identity) \\
    \frac{1}{2}\textrm{E}[
(\hat{z}^\ell)^2]  & (ReLU)
  \end{cases}
 \quad=a.
\end{eqnarray}
We then obtain
\begin{equation}
\label{simplifiedvariance}
\textrm{Var}[
{z}^\ell]=
an\textrm{Var}[
w]
+
\textrm{Var}[{z}^{\ell-1}]
=\cdots=
an\ell\textrm{Var}[
w]
+
\textrm{Var}[x]\approx
an\ell\textrm{Var}[
w]
.
\end{equation}

Let us consider backpropagation next.
Gradients are now given by deltas for $z$ as
\begin{equation}
\label{appendixbp}
\frac{\partial E}{\partial \boldsymbol{W}^\ell}
=
\boldsymbol{\delta}^\ell_z\otimes
f(\hat{\boldsymbol{z}}^{\ell-1}),\quad
\boldsymbol{\delta}^\ell_z
=
\frac{\partial E}{\partial \boldsymbol{z}^\ell}.
\end{equation}
Applying chain rule leads to the following backpropagation rule
\begin{eqnarray}
\delta^\ell_{z,j}
=
\sum_k
\frac{\partial E}{\partial {z}^{\ell+1}_{k}}
\frac{\partial {z}^{\ell+1}_{k}}{\partial z^\ell_j}
=\delta^{\ell+1}_{{z},j}
+\sum_{k,i}
f'(\hat{z}^{\ell}_i)w^{\ell+1}_{ki}
\frac{\partial \hat{z}^\ell_{i}}{\partial z^\ell_j}
\delta^{\ell+1}_{{z},k}.
\end{eqnarray}
Using (\ref{ACdeltaFormula}), we can rewrite it into the following form
\begin{eqnarray}
\delta^\ell_{z,j}
=\delta^{\ell+1}_{{z},j}
+\sum_k
f'(\hat{z}^{\ell}_j)w^{\ell+1}_{kj}\left(
\frac{1}{\sigma^\ell_j}\left(1-\frac{1}{N}\right)
-
\frac{1}{\left( \sigma^\ell_j \right)^3N}
\left(z^\ell_j-\mu^\ell_j\right)^2
\right)
\delta^{\ell+1}_{{z},k}.
\end{eqnarray}

Let us take the limit $N\to\infty$ for simplicity.
The expectation value of $(\delta^\ell_z)^2$ is then
\begin{equation}
\textrm{Var}[
\delta^\ell_{z}]
=
\left(
1+
a\frac{n \textrm{Var}[w]}{(\sigma^\ell)^2}
\right)
\textrm{Var}[\delta^{\ell+1}_{z}]
=\left(
1+
\frac{1}{\ell}
\right)
\textrm{Var}[\delta^{\ell+1}_{z}]
\end{equation}
To obtain the last equality, we use (\ref{simplifiedvariance}).
This formula immediately leads to the following evolution equation
\begin{equation}
\textrm{Var}[
\delta^\ell_{z}]
=
\textrm{Var}[\delta^{L}_{z}]
\prod_{\ell'=\ell}^L
\left(
1+
\frac{1}{\ell'}
\right)
=\frac{L}{\ell}\textrm{Var}[\delta^{L}_{z}],
\end{equation}
for both choices of activation functions.

To compute variance of gradients by using (\ref{appendixbp}),
we need to evaluate that of
$f(\hat{\boldsymbol{z}}^{\ell-1})$.
For identity activation function, the variance of it is $1$.
For ReLU, the variance becomes
\begin{equation}
\textrm{Var}[f(\hat{\boldsymbol{z}}^{\ell-1})]
=
\textrm{E}[(f(\hat{\boldsymbol{z}}^{\ell-1}))^2]
-
(\textrm{E}[f(\hat{\boldsymbol{z}}^{\ell-1})])^2
=
\frac{1}{2}-
(\textrm{E}[f(\hat{\boldsymbol{z}}^{\ell-1})])^2.
\end{equation}
Let us approximate this numerical coefficient by $a$.
Then the final formula is
\begin{equation}
\textrm{Var}\left[
\frac{\partial E}{\partial {w}^\ell}\right]
\approx
a\frac{L}{\ell}\textrm{Var}[\delta^{L}_{z}].
\end{equation}
This behavior is completely the same as that in Section 3,
and therefore we can expect this property is universal for batch normalized ResNets.

\end{document}